\documentclass[10pt,twocolumn,letterpaper]{article}
\pdfoutput=1

\usepackage[pagenumbers]{cvpr} % To force page numbers, e.g. for an arXiv version
\usepackage[dvipsnames]{xcolor}

\definecolor{citecolor}{HTML}{0071BC}
\usepackage[pagebackref,breaklinks,colorlinks,citecolor=citecolor]{hyperref}
\usepackage{multirow}
\usepackage{ragged2e}
\usepackage{makecell}
\usepackage{xcolor}         % colors
\usepackage{colortbl}
\usepackage{amsmath}
\usepackage{amssymb}

\def\eg{\emph{e.g.}}
\def\ie{\emph{i.e.}}

\title{PointOBB: Learning Oriented Object Detection via Single Point Supervision}
\author{Junwei Luo$^{1}$, Xue Yang$^{2}$\thanks{Project lead.} , Yi Yu$^{3}$, Qingyun Li$^{4}$, Junchi Yan$^{5,2}$, Yansheng Li$^{1}$\thanks{Corresponding author.}\\
$^{1}$Wuhan University \quad $^{2}$Shanghai AI Laboratory   \\ $^{3}$Southeast University \quad
$^{4}$Harbin Institute of Technology \quad $^{5}$Shanghai Jiao Tong University \\
{\tt\small luojunwei@whu.edu.cn} \quad {\tt\small yangxue@pjlab.org.cn} \\
{\tt\small Code: \url{https://github.com/Luo-Z13/pointobb}}
}

\begin{document}
\maketitle

\begin{abstract}
Single point-supervised object detection is gaining attention due to its cost-effectiveness. However, existing approaches focus on generating horizontal bounding boxes (HBBs) while ignoring oriented bounding boxes (OBBs) commonly used for objects in aerial images. This paper proposes \textbf{PointOBB}, the first single \textbf{Point}-based \textbf{OBB} generation method, for oriented object detection. PointOBB operates through the collaborative utilization of three distinctive views: an original view, a resized view, and a rotated/flipped (rot/flp) view. Upon the original view, we leverage the resized and rot/flp views to build a scale augmentation module and an angle acquisition module, respectively. In the former module, a Scale-Sensitive Consistency (SSC) loss is designed to enhance the deep network's ability to perceive the object scale. For accurate object angle predictions, the latter module incorporates self-supervised learning to predict angles, which is associated with a scale-guided Dense-to-Sparse (DS) matching strategy for aggregating dense angles corresponding to sparse objects. The resized and rot/flp views are switched using a progressive multi-view switching strategy during training to achieve coupled optimization of scale and angle. Experimental results on the DIOR-R and DOTA-v1.0 datasets demonstrate that PointOBB achieves promising performance, and significantly outperforms potential point-supervised baselines.

\end{abstract}

\section{Introduction}
\label{sec:intro}

\begin{figure}[!t]
  \centering
  \includegraphics[width=\columnwidth]{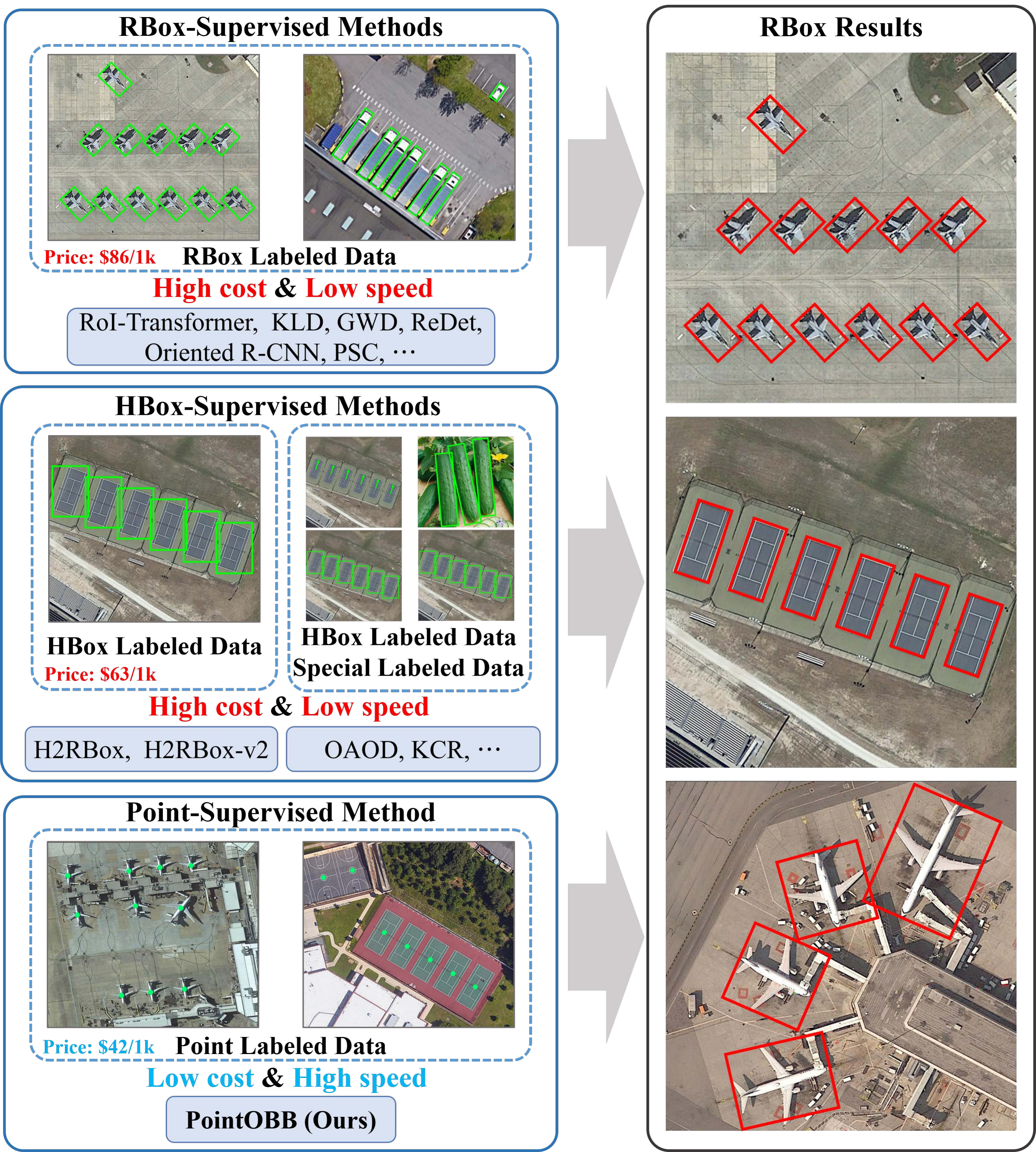}\\
  
  \caption{The main paradigmatic types of existing oriented object detection. Existing methods are primarily divided into rotated boxes (RBox)-supervised, horizontal boxes (HBox)-supervised, and Point-supervised approaches. Compared to RBox and HBox labels, point labels have lower costs and higher efficiency.}
  \vspace{-5pt}
  \label{fig:compare}
\end{figure}

Oriented object detection in aerial images aims to locate the objects of interest using oriented bounding boxes (OBBs) and identify the categories. There have been numerous outstanding research studies in this field~\cite{ding2019learning,yang2021r3det,han2021redet,xie2021oriented,yang2021learning,yang2021rethinking,li2023large}. However, manually annotating the fine-grained OBBs is time-consuming and expensive.

As to reducing the annotation cost, weakly-supervised horizontal object detection using image-level annotations has been well-developed~\cite{bilen2016weakly,tang2017multiple,tang2018pcl,wan2019min,chen2020slv,feng2022weakly,feng2023learning}, but these methods have limited performance in complex aerial scenes and cannot predict the orientation of objects. In recent years, there has been growing attention to weakly supervised oriented object detection. As illustrated in Fig.~\ref{fig:compare}, existing weakly-supervised methods employ coarser-grained annotations as weakly supervised signals to predict OBBs, \eg, horizontal box~\cite{sun2021oriented,yang2022h2rbox,zhu2023knowledge,yu2023h2rboxv2} annotations. However, box-based annotations are still inefficient and labor-intensive. Therefore, it is necessary to explore more cost-effective and efficient annotation forms.

Point-based annotation has recently gained attention in many tasks~\cite{ren2020ufo2,fan2022psps,cheng2022pointly,li2023point2mask}. In the field of object detection, the cost of point annotations is about 36.5\% lower than HBB and 104.8\% lower than OBB, and the efficiency of labeling point is significantly higher than both HBB and OBB\footnote{According to \url{https://cloud.google.com/ai-platform/data-labeling/pricing}, and point annotations are just 1.2x more time consuming than obtaining image-level annotations~\cite{bearman2016s, everingham2010pascal}.}. Therefore, single point-supervised oriented object detection seems more meaningful, and as far as we know, the relevant research is still blank.

From the bird’s-eye view, objects in aerial images showcase two specific characteristics: various spatial scales and arbitrary orientations. Considering plenty of tiny objects existing in aerial images~\cite{xia2018dota}, employing single point labels is undoubtedly more appropriate than multi point labels like~\cite{cheng2022pointly}. Existing single point-supervised object detection methods~\cite{papadopoulos2017click,chen2022p2b} follow the Multiple Instance Learning (MIL) fashion. This fashion optimizes through category labels, selecting the proposals with the highest confidence from proposal bags as the predicted boxes, thereby achieving perception of the object scale. However, the MIL fashion faces inherent instability in perceiving object boundaries, as it tends to focus on the most discriminative part of an object instead of its exact scale and boundary~\cite{tang2017multiple,tang2018pcl}. We identify two critical issues when extending this fashion to oriented object detection in aerial images: i) How to address the inconsistency in MIL to obtain more accurate scale representations? ii) How can we learn the object's orientation under single point supervision?

In this paper, we introduce \textbf{PointOBB}, the first single \textbf{Point}-based \textbf{OBB} generation framework. It achieves collaborative learning of both angle and scale by incorporating three unique views. Specifically, we construct a resized view and a rotated/flipped (rot/flp) view as enhanced views. To address the abovementioned two issues, we propose two modules: a scale enhancement module via the original and resized views, and an angle acquisition module relying on the original and rot/flp views. The core of the former module is a Scale-Sensitive Consistency (SSC) loss, which aims to address the aforementioned inconsistency between the proposal's confidence score and its scale accuracy. To obtain accurate object angle prediction, we propose a Dense-to-Sparse (DS) matching strategy associated with the self-supervised angle learning branch in the latter module. Additionally, the enhanced views are switched during training through a progressive multi-view switching strategy. The main contributions are as follows:

\begin{itemize}
  \item As far as we know, this paper proposes the first method, named PointOBB, to achieve oriented object detection under single point supervision. 
  \item The proposed method couples learning the object's scale and orientation through three distinctive views, guided via a progressive multi-view switching strategy.
  \item A SSC loss is designed to enhance the network's ability to perceive the object scale, and a scale-guided DS matching strategy is introduced to improve the accuracy of object angle prediction.
  \item Our method significantly outperforms existing competitive point-supervised alternatives on DIOR-R and DOTA-v1.0 datasets.
  
\end{itemize}

\section{Related works}
\label{sec:formatting}

\subsection{Fully-Supervised Oriented Object Detection}

Oriented object detection algorithms primarily focus on aerial objects~\cite{xu2020gliding}, multi-oriented scene texts~\cite{liao2018rotation,liu2018fots}, 3D-objects~\cite{yang2022detecting}, etc. Representative approaches include anchor-free detectors like Rotated FCOS~\cite{tian2019fcos}, and anchor-based detectors such as Rotated RetinaNet~\cite{lin2017focal}, RoI Transformer~\cite{ding2019learning}, Oriented R-CNN~\cite{xie2021oriented}, and ReDet~\cite{han2021redet}. Oriented RepPoints~\cite{li2022oriented} proposes a method for sample quality assessment and allocation based on adaptive points. Some methods like R$^3$Det~\cite{yang2021r3det} and S$^2$A-Net~\cite{han2021align} enhance the performance of the detectors through exploring feature alignment modules. To address the boundary discontinuity in angle regression, angle coders~\cite{yang2020arbitrary,yang2021dense,yu2023phase} transform the angle into boundary-free formats. Moreover, Gaussian-based losses like GWD~\cite{yang2021rethinking} and KLD~\cite{yang2021learning} analyze the nature of rotation representation and propose a Gaussian-based representation to improve performance.

\subsection{Weakly-Supervised Oriented Object Detection}
%-------------------------------------------------------------------------
\begin{figure*}[!tb]
	\begin{center}
		\includegraphics[width=0.99\linewidth]{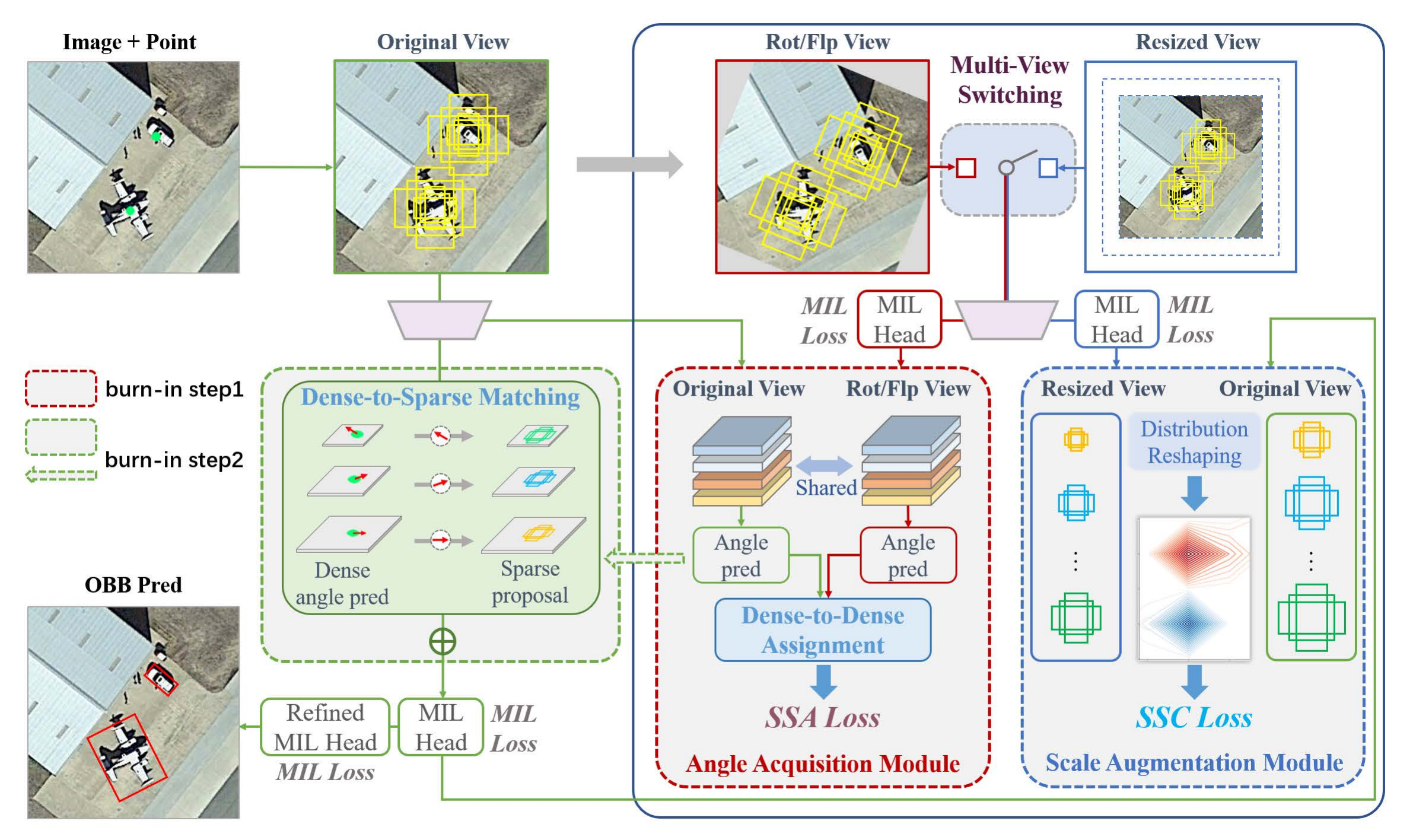}
	\end{center}
	\vspace{-2pt}
	\caption{The pipeline of the PointOBB. PointOBB contains three views in total. From these views, an angle acquisition module and a scale augmentation module are constructed. The scale augmentation module incorporates a Scale-Sensitive Consistency (SSC) loss to enhance scale perception ability. Within the angle acquisition module, a Dense-to-Dense sample assignment is designed for angle learning, optimized using the Self-Supervised Angle (SSA) loss. Additionally, a Dense-to-Sparse (DS) matching strategy is proposed to obtain more precise object angles. During the training, a progressive multi-view switching strategy is designed to switch the resized and rot/flp views, along with their corresponding modules.}
    \label{fig:pipeline}
\end{figure*}

Existing weakly-supervised oriented object detection approaches can be divided into Image-supervised and HBox-supervised methods. Furthermore, we explore the feasibility of Point-supervised methods.

\textbf{Image-Supervised.} For methods using image-level annotation, WSODet~\cite{tan2023wsodet} enhances the OICR~\cite{tang2017multiple} framework to predict HBBs. Then, it generates pseudo OBBs using contour features and predicted HBBs, followed by an Oriented RepPoints branch. However, image-supervised methods yield limited performance, and the generated OBBs are overly reliant on the quality of predicted HBBs.

\textbf{HBox-Supervised.} For methods using HBox-level annotation, although some methods (\eg, BoxInst-RBox~\cite{tian2021boxinst} and BoxLevelSet-RBox~\cite{li2022box}) use the HBox-Mask-RBox style to generate OBBs, involving segmentation is often more computationally costive, and the whole procedure can be time-consuming~\cite{yang2022h2rbox}. H2RBox~\cite{yang2022h2rbox} pioneeringly predicts the RBox directly from HBox annotations without redundant representations. It learns the angle from the geometry of circumscribed boxes and achieves remarkable performance. As a new version, H2RBox-v2~\cite{yu2023h2rboxv2} exploits the inherent symmetry of objects. However, such methods still require collecting a large number of bounding box annotations. Additionally, some studies utilize HBox and special forms of annotation. OAOD~\cite{iqbal2021leveraging} uses extra object angle, KCR~\cite{zhu2023knowledge} employs RBox-annotated source datasets with HBox-annotated target datasets. Sun et al.~\cite{sun2021oriented} combine HBox annotation and image rotation to align oriented objects horizontally or vertically. However, these specialized annotation forms lack universality.

\textbf{Point-Supervised.} Point-based annotations have been widely used in various tasks like object detection~\cite{papadopoulos2017click,ren2020ufo2,chen2021points,chen2022p2b,ying2023mapping,he2023learning}, panoptic segmentation~\cite{fan2022psps,li2023point2mask}, instance segmentation~\cite{bearman2016s,cheng2022pointly}, and so on~\cite{yu2022object,kirillov2023segment}. Due to its cost-effectiveness and efficiency, single-point supervised object detection has garnered attention. Click~\cite{papadopoulos2017click} makes an early exploration of point-supervised object detection. It proposes center-click annotations and incorporates them into MIL fashion. P2BNet~\cite{chen2022p2b} employs a coarse-to-fine strategy and adds negative samples to improve the quality of predictions. However, these methods only obtain horizontal boxes and disregard the inherent instability of MIL fashion. 

Based on the above, to finally obtain RBox through the point label, one potential method is the Point-to-Mask~\cite{li2023point2mask} approach, which entails finding the minimum circumscribed rectangle of the mask. Another possible method involves simply combining the Point-to-HBox and HBox-to-RBox approaches. In our experiments, these potential approaches are employed for comparison. Overall, no existing methods can directly achieve oriented object detection via single point supervision. This paper aims to fill this blank and provide a valuable starting point.

\section{Method}
\label{sec:method}

\subsection{Overall Framework}
\label{sec:3-1}

In existing image-supervised and point-supervised object detection methods, MIL-based paradigms demonstrate fundamental perceptual capabilities for object scale. Therefore, we employ the classic MIL fashion as the underlying network. The overall framework of our approach is illustrated in Fig.~\ref{fig:pipeline}. From the original view, initial proposal bags are generated via the point labels. Then angle predictions from the angle acquisition module are selected to align with these proposals through the DS matching strategy, equipping the horizontal proposals with orientation. From the generated rotated proposals, reliable ones are selected and refined via a MIL head and a refined MIL head to obtain the final OBB predictions. The acquired OBBs serve as pseudo labels for the final training of oriented object detectors.

The above pipeline employs three unique views in total. Based on the original view, the resized view is created by random scaling, while the rot/flp view is created by random rotating or vertical flipping. Through these three views, we construct a scale augmentation module and an angle acquisition module. The former module aims to enhance the network in perceiving object scales, while the latter is designed to learn object angles. These two views serve as enhanced views and are switched during the training process by the proposed progressive multi-view switching strategy.

\subsection{Progressive Multi-View Switching Strategy }
\label{sec:3-2}
To enable the network to gradually acquire discriminative capabilities for object scale and predictive abilities for object orientation, we design a progressive multi-view switching strategy to optimize our framework.

This strategy consists of three stages: i) In the \textbf{first stage}, we begin by constructing a resized view from the original view using a scale factor $\sigma$. Leveraging the scale-equivalence constraints between these two views, we design a scale augmentation module to enhance the network's accuracy in perceiving object scales, which will be elaborated in subsequent sections. ii) In the \textbf{second stage} (\ie, “burn-in step1” in Fig.~\ref{fig:pipeline}), the network has acquired fundamental perceptual abilities to the scale and boundary of objects. However, the orientation information is still lacking. Therefore, we switch the resized view to a rot/flp one to construct an angle acquisition module with the original view. This module utilizes the dense-to-dense sample assignment for self-supervised angle learning. iii) In the \textbf{third stage} (\ie, “burn-in step2” in Fig.~\ref{fig:pipeline}), the network has been able to obtain accurate angles from the dense feature. Utilizing the proposed DS matching strategy, we align the dense angle predictions with the sparse proposals by leveraging neighboring receptive fields to obtain the object orientation.

\subsection{Scale Augmentation Module}
\label{sec:3-3}
Objects in aerial images exhibit significant scale variation. Therefore, under the MIL paradigm, the inconsistency between the confidence scores of the predicted boxes and the actual positional accuracy will be further exacerbated. To address this issue, we introduce the scale augmentation module, which is centered around the design of Scale-Sensitive Consistency (SSC) loss.

In an ideal situation, the predicted size for the same object — the scale of the proposal with the highest confidence score — should be consistent across views with varying resolutions. Guided by this criterion, the SSC loss aims to minimize the disparity in distributions of predicted scores between the original and the resized views.

\begin{figure}[!t]
  \centering
  \includegraphics[width=\columnwidth]{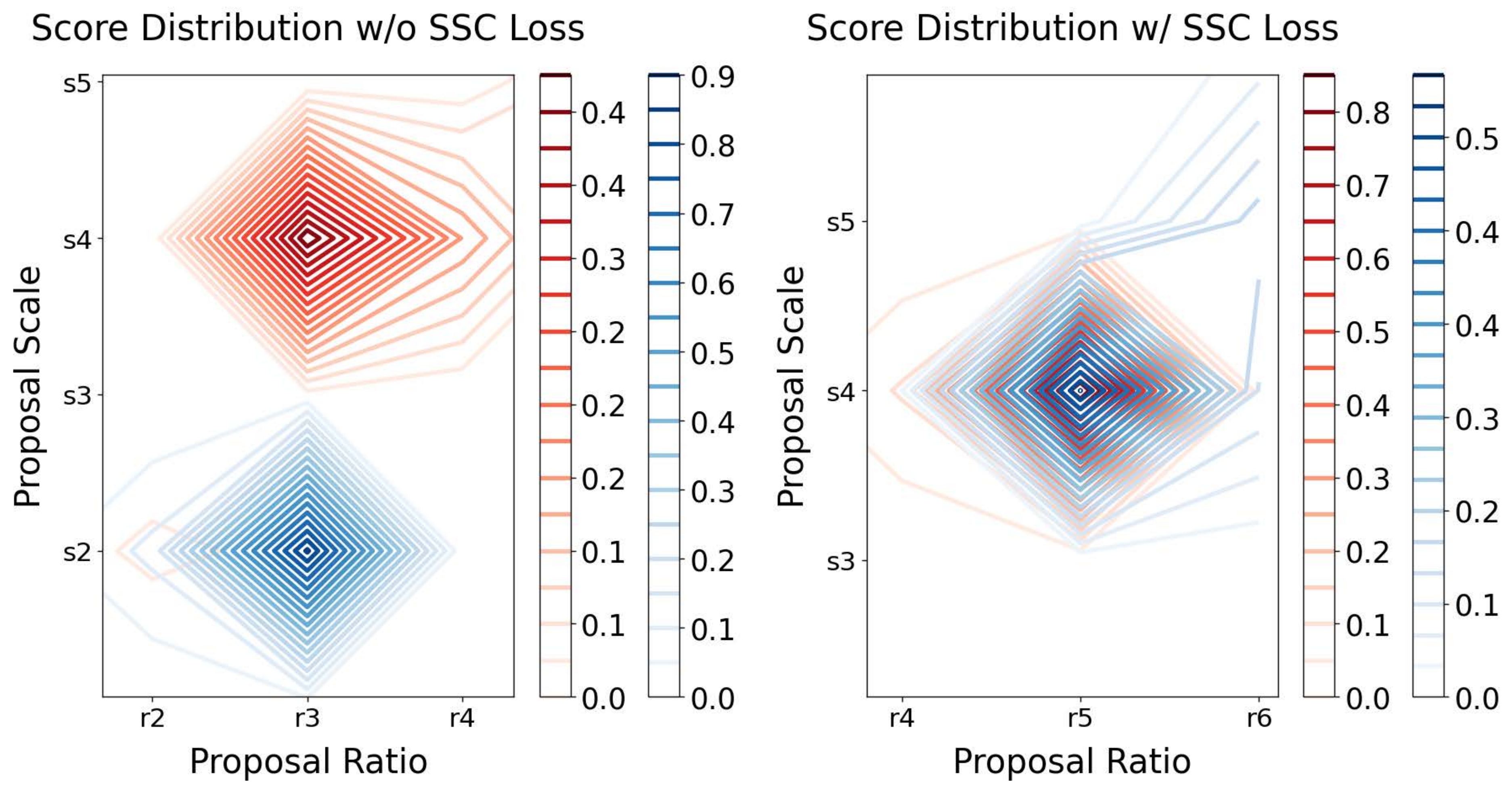}\\
  \vspace{-2pt}
  \caption{Distribution of the same proposal group's instance scores before and after applying the proposed SSC loss. We apply a local magnification of the coordinates for better visibility, and the scores are depicted on the right-hand side. The \textcolor{red}{red lines} and the \textcolor{blue}{blue lines} represent scores from the original view and resized view, respectively. It can be seen that the SSC loss effectively narrows the gap between these two score distributions.}
  \label{fig:Distribution}
\end{figure}

Given the original view's proposal bags $B_o$, we get the corresponding resized proposal bags $B_d$ from the resized view. The output scores (\ie, class scores and instance scores) of $B_o$ and $B_d$ are obtained by the dual-stream branches within the classical MIL head. To the $i$-th proposal bag, the scores derived from the original and resized views are denoted as $S^{cls}_{i_o}, S^{ins}_{i_o}$ and $S^{cls}_{i_d}, S^{ins}_{i_d}$, respectively. These sets of scores have been processed through an activation function (\eg, softmax), and their dimensions are $\mathbb{R}^{N\times C}$, where $N$ indicates the number of proposals in this bag and $C$ indicates the number of categories. To ensure scale equivalence among the outputs of different views, we first reshape the proposals' score distributions by the scale. Specifically, we define a set of basic scales $ \left \{ s_1,s_2,...,s_G \right \} $, where $G$ is the total number of the basic scales. The dimension of the output scores is reshaped from $\mathbb{R}^{N\times C}$ to $\mathbb{R}^{G\times K}$, where $K$ represents the number of scale-independent variables, such as aspect ratio and category. After reshaping the score distributions scale-wise from different views, we employ cosine similarity to measure their consistency:
\begin{align}
sim^{ins}_{m,g} &= 1- \frac{[{S^{ins}_{i_o}]_{m,g} \cdot [S^{ins}_{i_d}]_{m,g}}}{{\|[S^{ins}_{i_o}]_{m,g}\| \cdot \|[S^{ins}_{i_d}]_{m,g}\|}}, \\
sim^{cls}_{m,g} &= 1- \frac{[{S^{cls}_{i_o}]_{m,g} \cdot [S^{cls}_{i_d}]_{m,g}}}{{\|[S^{cls}_{i_o}]_{m,g}\| \cdot \|[S^{cls}_{i_d}]_{m,g}\|}},
\end{align}
where $m$ represents the $m$-th point label, and $g$ represents the $g$-th group of proposals after being grouped based on the basic scales. With the similarity measurements, the overall SSC loss is formulated as:
\begin{equation}
\mathcal{L}_{SSC} = \sum_{m=1}^{M} \sum_{g=1}^{G} \left \{  \omega_1 \ell_s(sim^{ins}_{m,g},0) + \omega_2 \ell_s(sim^{cls}_{m,g},0) \right \},
\end{equation}
where $M$ is the number of point labels, $\ell_s$ is SmoothL1 loss, $\omega_1$ and $\omega_2$ are weights and set to 2.0 and 1.0, respectively. Utilizing the SSC loss, the MIL network aligns the distribution of scores for proposals from the same label across different views, as depicted in Fig.~\ref{fig:Distribution}. This alignment helps mitigate inconsistencies between confidence scores and positional precision, enhancing the accuracy of perceiving object scale.

\subsection{Angle Acquisition Module}
\label{sec:3-4}

To learn the orientation under the absence of angle supervision, we begin by considering the inherent symmetry of objects. Previous exploration has studied symmetry under HBox supervision~\cite{yu2023h2rboxv2}. We discover that symmetry-based self-supervised learning has robustness to annotation noise. This indicates that even from a single point, accurate angle prediction is potential.

However, the absence of scale information makes it challenging to apply common sample assignment strategies in angle learning. Therefore, it is crucial to select appropriate samples for angle learning and match the angle prediction with corresponding objects to obtain final OBBs. To achieve this, we construct the angle acquisition module via the rot/flp view, with the proportion between rotate and flip set to 95:5 as ~\cite{yu2023h2rboxv2}. Angle learning is accomplished through dense-to-dense sample assignment, and the dense-to-sparse matching strategy realizes angle matching.

\textbf{Dense-to-Dense Assignment.} 
As described in Sec.~\ref{sec:3-1}, in the second stage, an angle acquisition module is built upon the rot/flp view, and it contains a self-supervised angle branch for angle learning. Both the views are sent into parameter-shared feature extractors (\eg, ResNet50~\cite{he2016deep} and FPN~\cite{lin2017feature}) to obtain dense feature pyramid features. As scale information is lacking, we select grid points on all feature levels on the center area around the ground-truth points as positive samples. For positive samples corresponding to the same point label in a level, we obtain the average of their predicted angles as the prediction value.

\textbf{Dense-to-Sparse Matching.} To match the dense feature-based angle predictions and the sparse feature-based proposals, simply searching for the nearest grid points to the center of a proposal is inappropriate. Due to the potential disparity between the grid points' receptive field and the object proposals' scale, the angle predictions may not be based on the actual object region. We perform a hierarchical pairing to consistently match the receptive fields involved in angle prediction with the scale of proposals. Assuming the feature pyramid is $lvls_{fea} = \left \{ p_1,p_2,...p_P \right \} $ with $P$ levels, we also classify the proposal into several levels based on the scale, $lvls_{prop} = \log_2 \left( \frac{{\sqrt{{w \cdot h}}}}{basescale} + 1e-6 \right)$, where $basescale$ is a pre-set scale parameter, $w$ and $h$ represent the width and height of the proposal, respectively. For each corresponding pair of levels $lvls_{fea}$ and $lvls_{prop}$, we utilize the average angle predictions from the central region as the orientation of the proposal, as shown in Fig.~\ref{fig:pipeline}, thereby aggregating the dense angle predictions for the sparse objects.

\textbf{SSA Loss.} According to the affine transformation relationship between the original view and the rot/flp view, object angles can be learned through the Self-Supervised Angle (SSA) loss. Assuming that the enhanced view is generated via random rotation with an angle $\theta'$, the angle predictions from both the original and rotated view are supposed to satisfy the same rotation relationship. If the enhanced view is generated via vertical flip, the angle predictions should satisfy the differences $k\pi$, where $k$ is an integer to keep the results in the same cycle. The loss between the outputs of two views can be represented as:
\begin{align}
\begin{aligned}
   & \begin{cases} 
   \mathcal{L}_{\text{rot}} = \min_{k \in \mathbb{Z}} \sum_{p=1}^{P} \ell_{\text{angle}}(\theta_{\text{rot}}^p - \theta^p, k \pi + \theta') \\
   \mathcal{L}_{\text{flp}} = \min_{k \in \mathbb{Z}} \sum_{p=1}^{P} \ell_{\text{angle}}(\theta_{\text{flp}}^p + \theta^p, k\pi)
   \end{cases} ,
\end{aligned}
\end{align}
where $\ell_{angle}$ is SmoothL1 loss, $\theta^p$, $\theta_{flp}^p$, and $\theta_{rot}^p$ represent the angle predictions from the $p$-th level of the feature pyramids from the original view, rot/flp view by flip, and rot/flp view by rotation, respectively. The SSA loss is defined as:
\begin{equation}
    \mathcal{L}_{SSA}= \mathcal{L}_{\text{rot}} + \mathcal{L}_{\text{flp}}.
\end{equation}

\subsection{Optimization with Overall MIL Loss}
\label{sec:3-5}

\begin{table*}[tb!]
    \centering
    \resizebox{1.0\textwidth}{!}{
    \fontsize{12pt}{12.0pt}\selectfont
        \renewcommand\arraystretch{1.6}
        \setlength{\tabcolsep}{0.8mm}
        \begin{tabular}{l|cccccccccccccccccccc|cc}
        \toprule
        \textbf{Method} & \textbf{\underline{APL}} & \textbf{APO} & \textbf{\underline{BF}}  &  \textbf{\underline{BC}} & \textbf{BR} & \textbf{CH} & \textbf{ESA} & \textbf{ETS} & \textbf{DAM} & \textbf{GF} & \textbf{\underline{GTF}} &  \textbf{\underline{HA}} & \textbf{OP} & \textbf{\underline{SH}} & \textbf{STA} & \textbf{STO} & \textbf{\underline{TC}} &  \textbf{TS} &  \textbf{\underline{VE}} & \textbf{WM} & \textbf{\underline{8-mAP$_{50}$}} & \textbf{mAP$_{50}$} \\ \hline
        \multicolumn{22}{l}{\textbf{RBox-supervised:}} \\ \hline
        Rotated RetinaNet~\cite{lin2017focal}& 58.9 & 19.8 & 73.1 & 81.3 & 17.0 & 72.6 & 68 & 47.3 & 20.7 & 74.0 & 73.9 & 32.5 & 32.4 & 75.1 & 67.2 & 58.9 & 81.0 & 44.5 & 38.3 & 62.6 & 64.26 & 54.96 \\
        Rotated FCOS~\cite{tian2019fcos} & 61.4 & 38.7 & 74.3 & 81.1 & 30.9 & 72.0 & 74.1 & 62.0 & 25.3 & 69.7 & 79.0 & 32.8 & 48.5 & 80.0 & 63.9 & 68.2 & 81.4 & 46.4 & 42.7 & 64.4 & 66.59 & 59.83 \\ 
        Oriented R-CNN~\cite{xie2021oriented}  & 63.1 & 34.0 & 79.1 & 87.6 & 41.2 & 72.6 & 76.6 & 65.0 & 26.9 & 69.4 & 82.8 & 40.7 & 55.9 & 81.1 & 72.9 & 62.7 & 81.4 & 53.6 & 43.2 & 65.6 &\textbf{69.88}&\textbf{62.80}\\ \hline
        \multicolumn{22}{l}{\textbf{Image-supervised:}} \\ \hline
        WSODet~\cite{tan2023wsodet}  & 20.7 & 29.0 & 63.2 & 67.3 & 0.2 & 65.5 & 0.4 & 0.1 & 0.3 & 49.0 & 28.9 & 0.3 & 1.5 & 1.2 & 53.4 & 16.4 & 40.0 & 0.1 & 6.1 & 0.1 & 28.46 & 22.20\\ \hline
        \multicolumn{22}{l}{\textbf{HBox-supervised:}} \\ \hline
        H2RBox~\cite{yang2022h2rbox} &68.1 & 13.0 & 75.0 & 85.4 & 19.4 & 72.1 & 64.4 & 60.0 & 23.6 & 68.9 & 78.4 & 34.7 & 44.2 & 79.3 & 65.2 & 69.1 & 81.5 & 53.0 & 40.0 & 61.5 & \textbf{67.80} & \textbf{57.80}\\
        H2RBox-v2~\cite{yu2023h2rboxv2} &67.2 & 37.7 & 55.6 & 80.8 & 29.3 & 66.8 & 76.1 & 58.4 & 26.4 & 53.9 & 80.3 & 25.3 & 48.9 & 78.8 & 67.6 & 62.4 & 82.5 & 49.7 & 42.0 & 63.1 & 64.06 & 57.64 \\ \hline
        \multicolumn{22}{l}{\textbf{Point-supervised:}} \\ \hline
        Point2Mask-RBox~\cite{li2023point2mask} & 15.6 & 0.1 & 50.6 & 25.4 &  4.2 & 50.9 & 23.8 & 17.5 & 8.1 & 0.8 & 9.6 & 1.4 & 15.3 & 1.6 & 5.8 & 6.3 & 17.9 & 7.1 & 4.5 & 9.2 & 15.25 & 13.77 \\
        P2BNet~\cite{chen2022p2b} + H2RBox~\cite{yang2022h2rbox}  & 52.7 & 0.1 & 60.6 & 80.0 & 0.1 & 22.6 & 11.5 & 5.2 & 0.7 & 0.2 & 42.8 & 2.8 & 0.2 & 25.1 & 8.6 & 29.1 & 69.8 & 9.6 & 7.4 & 22.6 & 42.65& 22.59\\
        P2BNet~\cite{chen2022p2b} + H2RBox-v2~\cite{yu2023h2rboxv2}  & 51.6 & 3.0 & 65.2 & 78.3 & 0.1 & 8.1 & 7.6 & 6.3 & 0.8 & 0.3 & 44.9 & 2.3 & 0.1 & 35.9 & 9.3 & 39.2 & 79.0 & 8.8 & 10.3 & 21.3 & 45.94 & 23.61\\
        \rowcolor{gray!20} Ours (Rotated FCOS)  & 58.4 & 17.1 & 70.7 & 77.7 & 0.1 & 70.3 & 64.7 & 4.5 & 7.2 & 0.8 & 74.2 & 9.9 & 9.1 & 69.0 & 38.2 & 49.8 & 46.1 & 16.8 & 32.4 & 29.6 & 54.80 & 37.31 \\
        \rowcolor{gray!20} Ours (Oriented R-CNN)  & 58.2 & 15.3 & 70.5 & 78.6 & 0.1 & 72.2 & 69.6 & 1.8 & 3.7 & 0.3 & 77.3 & 16.7 & 4.0 & 79.2 & 39.6 & 51.7 & 44.9 & 16.8 & 33.6 & 27.7 &\textbf{57.38}&\textbf{38.08}\\ \bottomrule
        \end{tabular}
    }
    \caption{Accuracy on the DIOR-R testing set. The categories in DIOR-R include Airplane (APL), Airport (APO), Baseball Field (BF), Basketball Court (BC), Bridge (BR), Chimney (CH), Expressway Service Area (ESA), Expressway Toll Station (ETS), Dam (DAM), Golf Field (GF), Ground Track Field (GTF), Harbor (HA), Overpass (OP), Ship (SH), Stadium (STA), Storage Tank (STO), Tennis Court (TC), Train Station (TS), Vehicle (VE) and Windmill (WM). \textbf{\underline{8-mAP$_{50}$}} means the mAP of APL, BF, BC, GTF, HA, SH, TC, and VE, which are representative categories in remote sensing. “-RBox” means the minimum rectangle operation is performed on the Mask to obtain RBox.}
    \label{table:dior}
\end{table*}

\begin{table*}[tb!]
    \centering
    \resizebox{1.0\textwidth}{!}{
    \fontsize{12pt}{12.0pt}\selectfont
    \renewcommand\arraystretch{1.6}
    \setlength{\tabcolsep}{1.5mm}
    \setlength{\abovecaptionskip}{2mm} 
        \begin{tabular}{l| ccccccccccccccc|cc}
                \toprule
            \textbf{Method} & \textbf{\underline{PL}}    & \textbf{\underline{BD}}    & \textbf{BR}    & \textbf{\underline{GTF}}   & \textbf{\underline{SV}}    & \textbf{\underline{LV}}    & \textbf{\underline{SH}}    & \textbf{TC}    & \textbf{BC}    & \textbf{ST}    & \textbf{SBF}   & \textbf{RA}    & \textbf{\underline{HA}}    & \textbf{SP}    & \textbf{HC}    & \textbf{\underline{7-mAP$_{50}$}} & \textbf{mAP$_{50}$}  \\ \hline
            \multicolumn{18}{l}{\textbf{RBox-supervised:}} \\ \hline
            Rotated RetinaNet~\cite{lin2017focal} & 88.7 & 77.6 & 38.8 & 58.2 & 74.6 & 71.6 & 79.1 & 88.0 & 80.2 & 72.3 & 52.8 & 58.6  & 62.6 & 67.7 & 59.6  &73.19 &68.69\\
            Rotated FCOS~\cite{tian2019fcos} & 88.4 & 76.8 & 45.0 & 59.2 & 79.2 & 79.0 & 86.9 & 88.1 & 76.6 & 78.8 & 58.6 & 57.5 & 69.3 & 72.4 & 53.5  &\textbf{76.96} &\textbf{71.28} \\ \hline
            \multicolumn{18}{l}{\textbf{HBox-supervised:}} \\ \hline
            Sun et al.~\cite{sun2021oriented} &51.5  & 38.7  & 16.1  & 36.8  & 29.8  & 19.2  & 23.4  & 83.9  & 50.6  & 80.0    & 18.9  & 50.2  & 25.6  & 28.7  & 25.5 &32.14 &38.60 \\
            H2RBox~\cite{yang2022h2rbox} &88.5 & 73.5 & 48.8 & 56.9 & 77.5 & 65.4 & 77.9 & 88.9 & 81.2 & 79.2 & 55.3 & 59.9  & 52.4 & 57.6 & 45.3 & 70.29 & 67.21 \\
            H2RBox-v2~\cite{yu2023h2rboxv2} &89.0 & 74.4 & 51.0 & 60.5 & 79.8 & 75.3 & 86.9 & 90.9 & 86.1  & 85.0 & 59.2 & 63.2  & 65.2  & 71.6 & 49.7 &\textbf{75.88} &\textbf{72.52} \\ \hline
            \multicolumn{18}{l}{\textbf{Point-supervised:}} \\ \hline
            P2BNet~\cite{chen2022p2b} + H2RBox~\cite{yang2022h2rbox} & 24.7 & 35.9 & 7.0  & 27.9 & 3.3  & 12.1 & 17.5 & 17.5 & 0.8   & 34.0 & 6.3  & 49.6 & 11.6  & 27.2 & 18.8 & 19.02 &19.63 \\
            P2BNet~\cite{chen2022p2b} + H2RBox-v2~\cite{yu2023h2rboxv2} &11.0 & 44.8 & 14.9  & 15.4 & 36.8 & 16.7 & 27.8  & 12.1 & 1.8  & 31.2 & 3.4  & 50.6 & 12.6 & 36.7 & 12.5 & 23.58 &21.87 \\
            \rowcolor{gray!20} Ours (Rotated FCOS) & 26.1  & 65.7 & 9.1  & 59.4 & 65.8  & 34.9  & 29.8 & 0.5   & 2.3   & 16.7  & 0.6  & 49.04 & 21.8  & 41.0 & 36.7 & 43.35 & 30.08 \\
            \rowcolor{gray!20} Ours (Oriented R-CNN) & 28.3 & 70.7 & 1.5  & 64.9 & 68.8 & 46.8 & 33.9 & 9.1  & 10.0  & 20.1 & 0.2  & 47.0 & 29.7 & 38.2 & 30.6 & \textbf{49.01} & \textbf{33.31} \\ \bottomrule
        \end{tabular}
        }
    \caption{Accuracy results on the DOTA-v1.0 testing set. \textbf{7-mAP$_{50}$} means the mAP of 7 representative categories: Plane (PL), Baseball Diamond (BD), Ground Track Field (GTF), Small Vehicle (SV), Large Vehicle (LV), Ship (SH), and Harbor (HA).}
    \label{table:dota}
\end{table*}

In addition to the previously mentioned SSC and SSA loss, we introduce the MIL loss involved in our approach. As in Fig. \ref{fig:pipeline}, each constructed view undergoes the MIL head, and the output of the original view passes through an extra refined MIL head to obtain the final predictions. The corresponding loss can be summarized as follows. 

Specifically, each point label is associated with a corresponding proposal bag. Through the MIL head, the output instance scores and class scores of the $n$-th proposal in the $i$-th proposal bag is $\left (S^{ins}_{i,n}, S^{cls}_{i,n}\right )$. The output of the whole proposal bag can be denoted as $S_{i} = \sum_{n=1}^{N} S^{ins}_{i,n}\odot S^{cls}_{i,n}$, where $N$ is the number of proposals in this bag. The MIL losses for the original view, rot/flp view, and resized view are denoted as $\mathcal{L}_{MIL_{ori}}$, $\mathcal{L}_{MIL_{rfv}}$, and $\mathcal{L}_{MIL_{res}}$, respectively. These three losses follow the common form of the general MIL loss $\mathcal{L}_{MIL_{init}}$, which is defined as:
\begin{equation}
\begin{split}
    \mathcal{L}_{MIL_{init}}& = -\frac{1}{I} \sum_{i=1}^{I} \sum_{c=1}^{C} \Bigl\{  Q_{i,c}\log(S_{i,c}) \\
    & + (1-Q_{i,c})\bigl(1-\log(1-S_{i,c})\bigr) \Bigr\},
\end{split}
\end{equation}
where $I$ is the number of proposal bags in the batch, $C$ is the number of categories, $Q_{i,c}$ is the one-hot category label, and $S_{i,c}$ represents the score of $c$-th category in $S_{i}$. For the refined MIL head, the $\mathcal{L}_{MIL_{ref}}$ employs focal loss~\cite{lin2017focal} to calculate the classification loss between $Q_{i,c}$ and $S_{i,c}$. Therefore, considering the proposed progressive multi-view switching strategy, the overall MIL loss is defined as:
\begin{equation}
\begin{aligned}
  \mathcal{L}_{MIL} &= \mathcal{L}_{MIL_{ori}} +\mathcal{L}_{MIL_{ref}} 
  \\
  &+\alpha\mathcal{L}_{MIL_{rfv}}+ \beta \mathcal{L}_{MIL_{res}},
\label{eq:mil}
\end{aligned}
\end{equation}
where $\alpha$ is 1, $\beta$ is 0 in the first stage, and $\alpha$ is 0, $\beta$ is 1 in the second and third stages as depicted in Sec.~\ref{sec:3-2}. The overall loss of our framework is expressed as:
\begin{equation}
    \mathcal{L} = \mathcal{L}_{MIL} + \alpha \mathcal{L}_{SSC} + \beta \mathcal{L}_{SSA},
\end{equation}
where $\alpha$ and $\beta$ keep the same setting with Eq.~\ref{eq:mil}.

\section{Experiments}
\label{sec:experiment}

\begin{table}[tb!]
    \centering
    \fontsize{8.5pt}{12.0pt}\selectfont
    \setlength{\tabcolsep}{1mm}
    \setlength{\abovecaptionskip}{2mm} 
        \begin{tabular}{cc|ccc|ccc}
        \toprule
        \multicolumn{2}{c|}{\textbf{Module}} & \multicolumn{3}{c|}{\textbf{DIOR-R~\cite{cheng2022anchor}}} & \multicolumn{3}{c}{\textbf{DOTA-v1.0~\cite{xia2018dota}}} \\ \hline
            \textbf{SSC} & \textbf{DS} & \textbf{mIoU} & \textbf{8-mAP$_{50}$} & \textbf{mAP$_{50}$} & \textbf{mIoU}& \textbf{7-mAP$_{50}$}&\textbf{mAP$_{50}$} \\ \hline
            & &  47.95 & 47.40 &30.16& 40.26&43.65 &28.44 \\
            \checkmark & &  53.17 & 52.19 &36.39 & 43.92&47.03 &32.98\\
            & \checkmark &  50.78 & 49.89&31.96 & 42.54&44.34 &30.63 \\
            \rowcolor{gray!20} \checkmark & \checkmark  & \textbf{56.08} & \textbf{54.80} &\textbf{37.31}& \textbf{45.35}&\textbf{49.01}&\textbf{33.31} \\ \bottomrule
        \end{tabular}
    \caption{The effects of Scale-Sensitive Consistent (SSC) loss and Dense-to-Sparse (DS) matching strategy in DIOR-R and DOTA-v1.0 datasets. We train Rotated FCOS on the DIOR-R testing set and Oriented R-CNN on the DOTA-v1.0 testing set.}
    \label{table:Ablation}
\end{table}

\begin{figure*}[!tb]
	\begin{center}
		\includegraphics[width=1.0\linewidth]{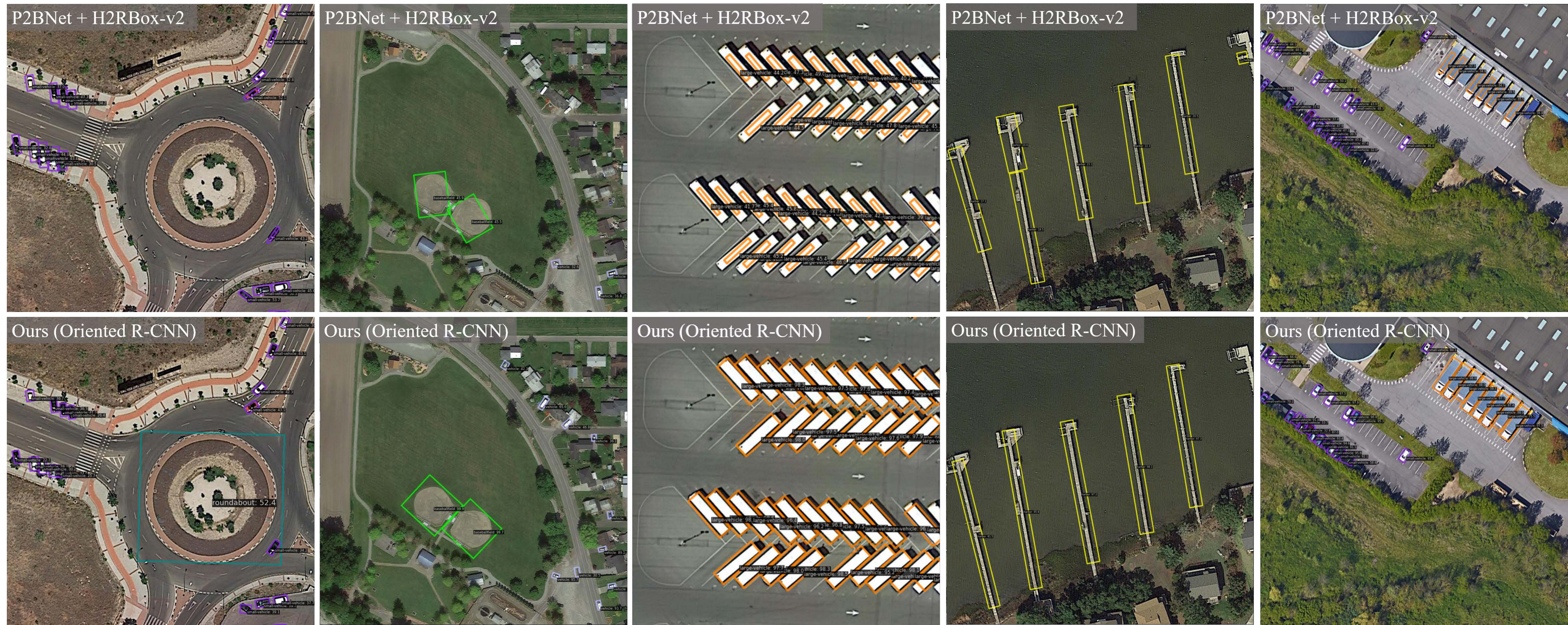}
	\end{center}
	\vspace{-8pt}
	\caption{The visual detection results. The first row showcases the results of the P2BNet~\cite{chen2022p2b}-H2RBox-v2~\cite{yu2023h2rboxv2} cascade pipeline. The second row showcases the results of the proposed PointOBB combined with Oriented R-CNN.}

    \label{fig:vis_result}
\end{figure*}

\subsection{Datasets and Implementation Details}
\label{sec:4-1}

\textbf{Dataset.} DIOR-R~\cite{cheng2022anchor} is an aerial image dataset with OBB annotation based on its previous version DIOR~\cite{li2020object}, which consists of 23,463 images, 20 categories, and 190,288 instances. DOTA-v1.0~\cite{xia2018dota} is a large-scale aerial images dataset for object detection, which contains 2,806 images, 15 categories, and 188,282 instances with both OBB and HBB annotations. Algorithms are trained on both the training and the validation set, and evaluated on the testing set.

\begin{table}[t]
    \fontsize{8.5pt}{12.0pt}\selectfont
    \setlength{\tabcolsep}{1.8mm}
    \setlength{\abovecaptionskip}{2.3mm}
    \centering
        \begin{tabular}{c c c c c c}
        \toprule
        \textbf{Original} & \textbf{Resized} & \textbf{Rot/Flp} & \textbf{mIoU} & \textbf{8-mAP$_{50}$} &\textbf{mAP$_{50}$} \\ \hline
        \checkmark &  &  & 52.90 & 53.45& 34.94\\
        \checkmark &  \checkmark &  & 54.44 & 54.73 & 36.26\\
        \checkmark &  &  \checkmark& 53.49 & 54.92& 35.80\\
        \rowcolor{gray!20} \checkmark & \checkmark & \checkmark & \textbf{56.08} & \textbf{57.38}& \textbf{38.08}\\ \bottomrule
       \end{tabular}
    \caption{Ablation studies of using MIL loss at different views on DIOR-R. mAP is reported by training the Oriented R-CNN.}
    \label{table:Ablation-MILLosstype}
\end{table}

\noindent\textbf{Single Point Annotation.}
To accurately reproduce the biases during manual annotation, we generate point labels based on the central region of the OBB labels. Specifically, we select a random point within a range comprising 10$\%$ in height and width relative to the OBB as the label. The effect of the range will be discussed in Sec.~\ref{sec:ablation}.

\noindent\textbf{Experiment Settings.}
The algorithms employed in our experiments are from two open-source pytorch-based algorithm libraries, MMRotate~\cite{zhou2022mmrotate} and MMDetection~\cite{chen2019mmdetection}. We follow the default setting in MMRotate, for the DOTA-v1.0 dataset, large-size images are cropped into 1,024 $\times$ 1,024 patches with a 200-pixel overlap. For the DIOR-R dataset, we keep the image size at the original size of 800$\times$800.

Experiments are performed on a server with 2 Tesla V100 GPUs and 16GB memory. We adopt the “2×” schedule for training all methods and adopt the “1×” schedule when using our generated pseudo OBBs to train RBox-supervised algorithms. The SGD optimizer is employed with a learning rate of 0.005, momentum of 0.9, and weight decay of 0.0001. A linear warm-up strategy is applied for the initial 500 iterations with a rate of 0.001 and batch size is 2. Apart from WSODet~\cite{tan2023wsodet} using VGG16~\cite{girshick2014rich}, all the listed models are configured based on the ResNet50~\cite{he2016deep} backbone which is first pre-trained on ImageNet~\cite{deng2009imagenet}. For the algorithms used for comparison, we follow their default settings. For our method, the $basescale$ is set to 56 and the resized view's scale factor $\sigma$ is randomly chosen between 0.5 and 1.5. The “burn-in step1” and “burn-in step2” are set to the 6th epoch and the 8th epoch, respectively. During the training, random flipping is employed as the only data augmentation technique.

\noindent\textbf{Evaluation Metric.} Mean Average Precision (mAP) is adopted as the primary metric to compare our methods with existing alternatives. To better assess the quality of pseudo-labels generated from points, we report the mean Intersection over Union (mIoU) between ground-truth boxes and predicted pseudo OBBs in the training set.

\subsection{Main Results}

The comparisons on the DIOR-R and DOTA-v1.0 datasets 
are shown in Tab.~\ref{table:dior} and Tab.~\ref{table:dota}. 

\noindent\textbf{Results on DIOR-R.} As shown in Tab.~\ref{table:dior}, PointOBB achieves AP$_{50}$ of 37.31$\%$ and 38.08$\%$ by training Rotated FCOS~\cite{tian2019fcos} and Oriented R-CNN~\cite{xie2021oriented}. For comparison, we employ state-of-the-art Point-to-HBox-to-RBox (\ie, P2BNet~\cite{chen2022p2b} + H2RBox-v2~\cite{yu2023h2rboxv2}) and Point-to-Mask-to-RBox (\ie, Point2Mask-RBox~\cite{li2023point2mask}) as plain two-stage approaches. Our results significantly outperform these two-stage approaches, exhibiting a minimum improvement of 14.47$\%$ in the mAP$_{50}$ metric. What's more, it is worth noting that our method results in performances of over \textbf{$85\%$} comparable to fully-supervised methods in eight major categories (\ie, the 8-mAP$_{50}$ metric), demonstrating the potential of point supervision.

\noindent\textbf{Results on DOTA-v1.0.} 
As shown in Tab.~\ref{table:dota}, our method achieves 33.31$\%$ in mAP$_{50}$ metric with Oriented R-CNN, surpassing the plain two-stage approach by 11.44$\%$.

The visual result is shown in Fig.~\ref{fig:vis_result}. The proposed PointOBB effectively addresses the issue of local focus MIL fashion and accurately predicts the object orientation.

\subsection{Ablation Studies}\label{sec:ablation}
In this section, we conduct several ablation studies to validate key parts of PointOBB. 

\begin{figure}[!tb]
	\centering
    \subfloat[SSA loss]{\includegraphics[width=0.23\textwidth]{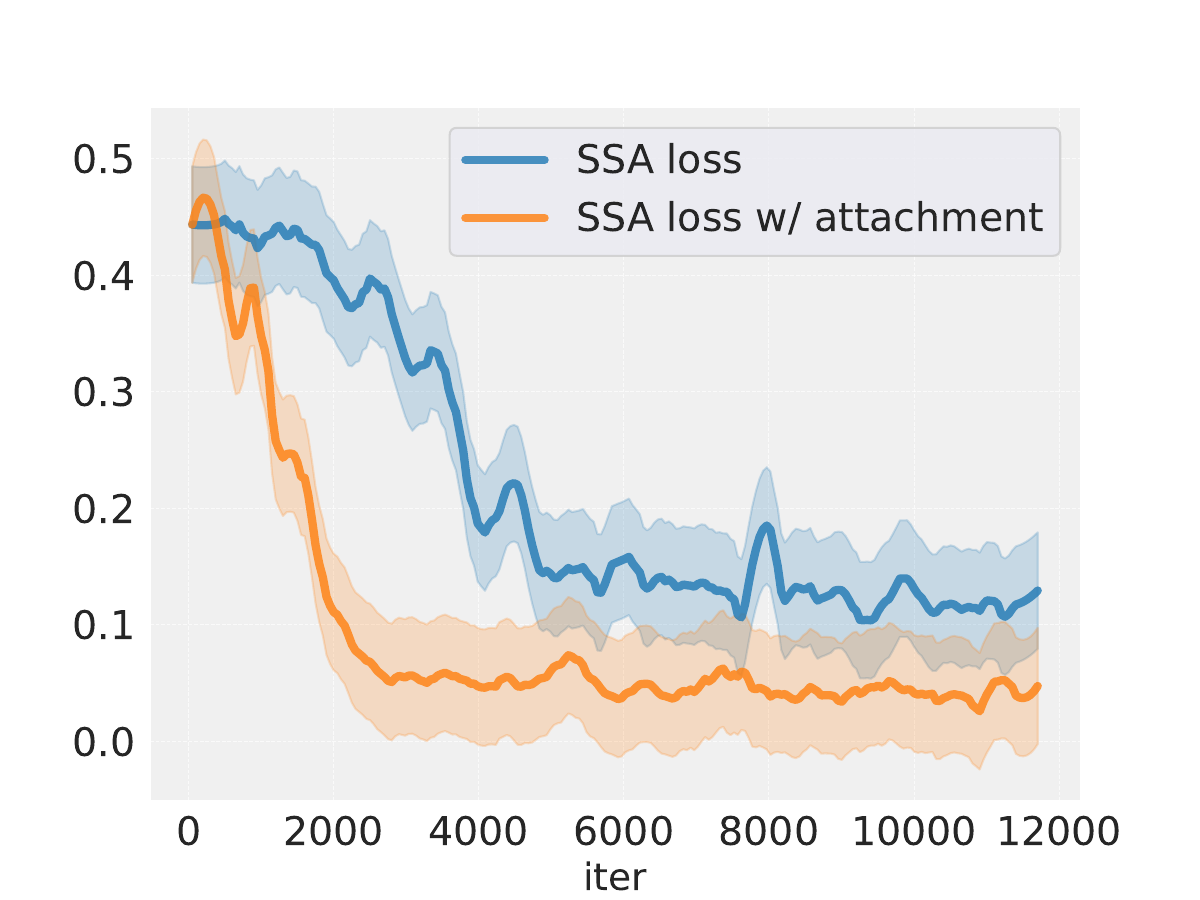}}
	\subfloat[MIL loss]{\includegraphics[width=0.23\textwidth]{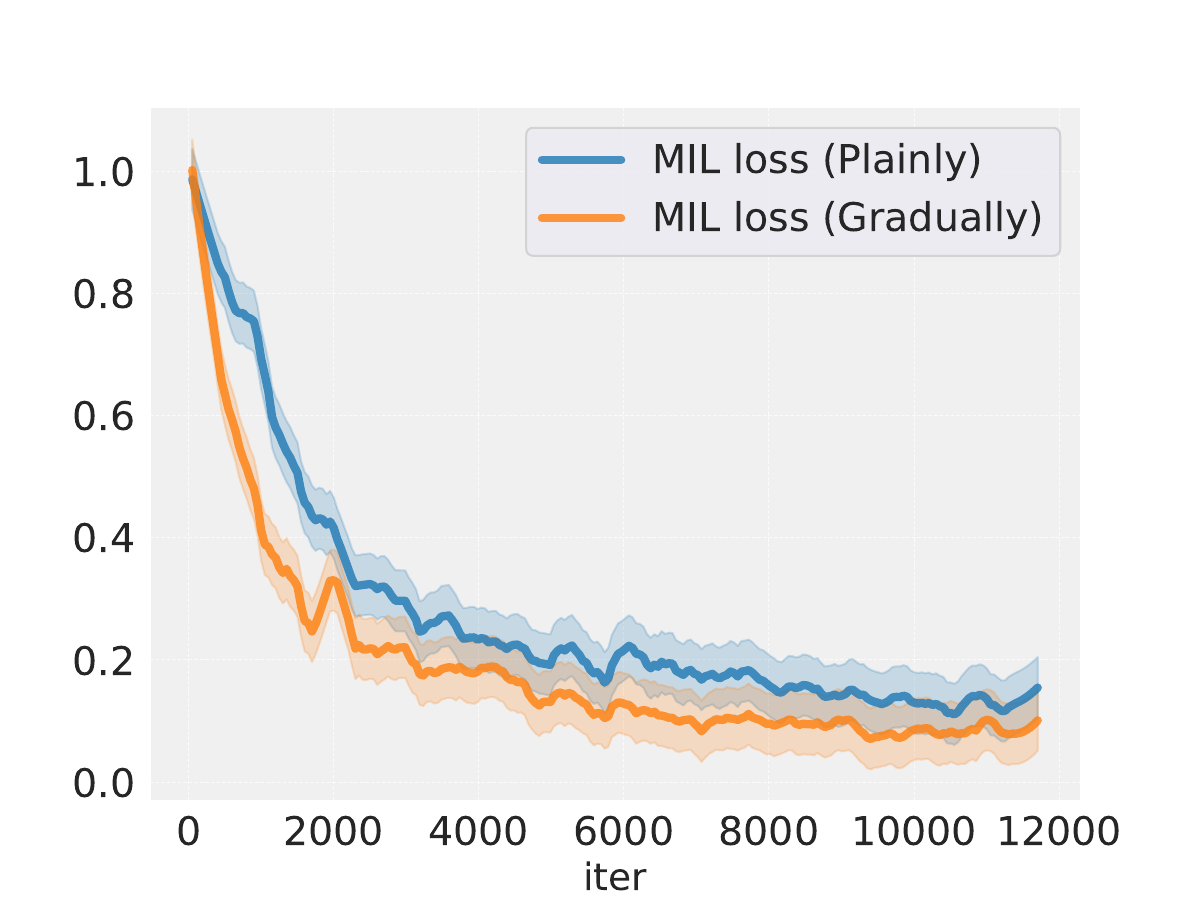}}	
	\caption{(a) indicates the SSA loss under different settings. In (b), ``MIL loss (Gradually)" means training the network with the proposed gradual strategy (\ie, in chronological order), and ``MIL loss (Plainly)" means training the angle branch from the beginning directly.}
	\label{fig:LossCurve}
\end{figure}

\begin{table}[tb!]
    \centering
    \fontsize{8.5pt}{12.0pt}\selectfont
    \setlength{\tabcolsep}{1.2mm}
    \setlength{\abovecaptionskip}{2mm} 
        \begin{tabular}{cccccc}
        \toprule
        \textbf{Plainly} & \textbf{Gradually} & \textbf{Attachment} & \textbf{mIoU} & \textbf{8-mAP$_{50}$}& \textbf{mAP$_{50}$} \\ \hline
            \checkmark &   &   & 34.92 & 36.13& 23.71\\
             \checkmark &    &  \checkmark  & 37.45 & 40.85&25.52\\
                 &  \checkmark  &   & 48.92 & 50.22 &32.23\\
                \rowcolor{gray!20} &   \checkmark  &  \checkmark  & \textbf{56.08} & \textbf{57.38} & \textbf{38.08}\\ \bottomrule
     \end{tabular}
     \caption{Ablation studies of the burn-in steps on DIOR-R. mAP is reported by training the Oriented R-CNN. ``Attachment" indicates attaching gradients between the angle branch and the other parts.}
    \label{table:BurninSteps}
\end{table}

\begin{table}[tb!]
    \centering
    \fontsize{8.5pt}{12.0pt}\selectfont
    \setlength{\tabcolsep}{1.5mm} 
    \setlength{\abovecaptionskip}{2mm} 
        \renewcommand\arraystretch{1.2}
        \begin{tabular}{l|cc>{\columncolor{gray!20}}c|c>{\columncolor{gray!20}}cc}
         \toprule
         \textbf{Parameter} & \multicolumn{3}{c|}{\textbf{Grouping Type}} & \multicolumn{3}{c}{\textbf{Point Range}} \\ \hline
         \textbf{Setting}   & proposal    & ratio    & scale   & 0$\%$     & 10$\%$   & 20$\%$   \\ \hline
         \textbf{mIoU}      & 43.95       & 54.38    & \textbf{56.08}   & 49.47   & \textbf{56.08}  & 54.21 \\
         \textbf{8-mAP$_{50}$} & 45.43  & 55.45 &  \textbf{57.38}   & 48.45 & \textbf{57.38} & 53.65 \\
         \textbf{mAP$_{50}$}  & 30.42  &  36.71 &  \textbf{38.08}   & 32.11 & \textbf{38.08} &35.78  \\ \bottomrule
\end{tabular}
    \caption{Ablation studies of the grouping type used in the SSC loss, and the generation range of point annotation on DIOR-R. mAP is reported by training the Oriented R-CNN.}
    \label{table:Ablation-Parameters}
\end{table}

\noindent\textbf{The effect of SSC and DS.}
Tab.~\ref{table:Ablation} studies the effects of the proposed two key designs: the SSC loss and the DS matching strategy. As shown in Tab.~\ref{table:Ablation}, both contribute to performance improvement. It indicates that the scale-consistency-based SSC loss has a significant effect, and the DS matching strategy also brings a boost by mitigating misalignment arising from unknown scales.

\noindent\textbf{The effect of different MIL losses.}
Tab.~\ref{table:Ablation-MILLosstype} illustrates the effects of applying MIL loss at different views. It can be observed that both enhanced views contribute to the improvement of accuracy, which may be because they represent different forms of data augmentation. 

\noindent\textbf{The effect of the angle branch's setting.} We explore the impact of the gradient backpropagation on the angle branch within the angle acquisition module. As shown in  Fig.~\ref{fig:LossCurve} (a) and Tab.~\ref{table:BurninSteps}, jointly optimizing the angle with the base network (``Attachment") leads to faster convergence and better accuracy. Emphasizing the necessity for collaborative optimization between the angle and scale.

\noindent\textbf{The effect of the burn-in steps.} Following the above ``Attachment" setting, we explore the impact of burn-in steps through MIL loss. We set burn-in steps to 0 as a ``Plainly" strategy, \ie, introducing angle learning at the beginning, and set our defined burn-in steps as ``Gradually" strategy. As shown in Fig.~\ref{fig:LossCurve} (b) and Tab.~\ref{table:BurninSteps}, the ``Plainly" strategy affects the initial optimization of the MIL network, resulting in performance degradation. In contrast, the "Gradually" strategy introduces the two burn-in steps when scale and angle learning essentially converge, effectively guiding the network to optimize in a progressive manner, thus yielding improvements.

\noindent\textbf{The effect of point range.}
The right part of Tab.~\ref{table:Ablation-Parameters} displays the impact of different ranges during point label generation. The introduction of appropriate noise proves to be advantageous compared to the center point label (\ie 0$\%$). We analyze that in the case of certain categories, the center point constrains the network's perceptual range, hindering the perception of object boundaries.

\noindent\textbf{The effect of grouping type in SSC loss.}
In the left part of Tab.~\ref{table:Ablation-Parameters}, we assess the impact of grouping scores based on scale, ratio, and proposal in the SSC loss. This demonstrates the effectiveness of scale-based grouping, aligning with the original intention of our design.

\section{Conclusion}
\label{sec:conclusion}
This paper introduces PointOBB, the first single point-based OBB generation framework for oriented object detection. PointOBB comprehensively learns the scale and orientation of objects through three distinctive views and is guided by a progressive multi-view switching strategy. Utilizing the three views, we construct a scale augmentation module and an angle acquisition module. The scale augmentation module enhances the network's capacity to perceive object scale by incorporating a Scale-Sensitive Consistency (SSC) loss. The angle acquisition module achieves self-supervised angle learning and further contributes to improving the accuracy of object angle prediction via Dense-to-Sparse (DS) matching. Our method outperforms the state-of-the-art alternatives on the DIOR-R and DOTA-V1.0 datasets. We hope this work can serve as a meaningful starting point for the development of single point-supervised oriented object detection.

\noindent\textbf{Future work.} PointOBB exhibits suboptimal accuracy for specific classes (e.g., BR, GF, and OP), possibly due to their relatively ambiguous definition of boundaries. This motivates further exploration and utilization of object features in aerial images, such as contextual features.

{
    \small
    \bibliographystyle{ieeenat_fullname}
    \bibliography{main}
}

\end{document}